\documentclass[lettersize,journal]{IEEEtran}
\usepackage{amsmath,amsfonts}
\usepackage{algorithmic}
\usepackage{algorithm}
\usepackage{array}
\usepackage[caption=false,font=normalsize,labelfont=sf,textfont=sf]{subfig}
\usepackage{textcomp}
\usepackage{stfloats}
\usepackage{url}
\usepackage{verbatim}
\usepackage{graphicx}
\usepackage{epsfig}
\usepackage{cite}
\usepackage{bbm}
\usepackage{hyperref}
\usepackage{booktabs}
\usepackage{framed,multirow}
\usepackage{color}
\usepackage{amsmath,amssymb,amsfonts}
\usepackage{array}
\usepackage{amssymb}  
\usepackage{pifont}   
\usepackage{textcomp}
\usepackage{stfloats}
\hypersetup{colorlinks,linkcolor=red,anchorcolor=blue,citecolor=blue}
\usepackage[numbers,sort&compress]{natbib}
\begin{document}

\title{An Evidential-enhanced Tri-Branch Consistency Learning Method for Semi-supervised Medical Image Segmentation}
\author{Zhenxi Zhang, Heng Zhou, Xiaoran Shi, Ran~Ran, Chunna Tian,~\IEEEmembership{Member,~IEEE}, Feng Zhou, \IEEEmembership{Member,~IEEE}
	\thanks{Manuscript received November 10 2023.} \thanks{Corresponding author: Feng Zhou.} 
	\thanks{Zhenxi Zhang, Xiaoran Shi and Feng Zhou are with the Ministry of Education, Key Laboratory of Electronic Information Counter-measure and Simulation, Xidian University, Xi’an 710071, China (e-mail:zhangzhenxi@xidian.edu.cn; xrshi@xidian.edu.cn; fzhou@mail.xidian.edu.cn).Chunna Tian and Heng Zhou are with School of Electronic Engineering, Xidian University, Xi’an 710071, China (e-mail:  chnatian@xidian.edu.cn; hengzhou@stu.xidian.edu.cn).
	Ran Ran is with Department of Medical Oncology, The First Affiliated
	Hospital of X’ıan Jiaotong University (email: ymyg123456@stu.xjtu.edu.cn)}}

\markboth{Journal of \LaTeX\ Class Files,~Vol.~, No.~, February~2023}%
{Shell \MakeLowercase{\textit{et al.}}: A Sample Article Using IEEEtran.cls for IEEE Journals}


\maketitle

\begin{abstract}
Semi-supervised segmentation presents a promising approach for large-scale medical image analysis, effectively reducing annotation burdens while achieving comparable performance. This methodology holds substantial potential for streamlining the segmentation process and enhancing its feasibility within clinical settings for translational investigations. While cross-supervised training, based on distinct co-training sub-networks, has become a prevalent paradigm for this task, addressing critical issues such as predication disagreement and label-noise suppression requires further attention and progress in cross-supervised training. In this paper, we introduce an Evidential Tri-Branch Consistency learning framework (ETC-Net) for semi-supervised medical image segmentation. ETC-Net employs three branches: an evidential conservative branch, an evidential progressive branch, and an evidential fusion branch. The first two branches exhibit complementary characteristics, allowing them to address prediction diversity and enhance training stability. We also integrate uncertainty estimation from the evidential learning into cross-supervised training, mitigating the negative impact of erroneous supervision signals. Additionally, the evidential fusion branch capitalizes on the complementary attributes of the first two branches and leverages an evidence-based Dempster-Shafer fusion strategy, supervised by more reliable and accurate pseudo-labels of unlabeled data. Extensive experiments conducted on LA, Pancreas-CT, and ACDC datasets demonstrate that ETC-Net surpasses other state-of-the-art methods for semi-supervised segmentation. The code will be made available in the near future at https://github.com/Medsemiseg.
\end{abstract}


\label{sec:introduction}
Medical image segmentation is a crucial task in various clinical applications \cite{wang2022disentangled,lin2022ds}, facilitating the identification of anatomical structures and pathological regions \cite{painchaud2020cardiac}. Deep learning algorithms \cite{xu2023cross,zhao2019deep} have demonstrated remarkable potential in automating and enhancing the precision of medical image segmentation~\cite{feng2020cpfnet}. While supervised deep learning methods rely on extensive labeled datasets for effective training, obtaining fully annotated medical images is hindered by the considerable manual workload of human experts, especially in a large-scale setting. To address this challenge, semi-supervised learning techniques \cite{cheplygina2019not} have emerged, harnessing both labeled and unlabeled data to achieve segmentation performance comparable to fully supervised methods using exclusively labeled data. Among the approaches in semi-supervised learning for image segmentation, two widely investigated categories are consistency-based and pseudo-label-based techniques. Consistency regularization is grounded under the smoothness assumption, which asserts that predictions from both labeled and unlabeled data should demonstrate consistency under various perturbations. Methods such as Mean Teacher~\cite{tarvainen2017mean} enhances consistent predictions between a teacher model and a student model, where the teacher network maintains a moving average of the student network's parameters as training progresses. Luo et al. \cite{luo2021efficient} propose an uncertainty rectified pyramid consistency method (URPC) which introduces a consistency constraint across multi-scale predictions by incorporating uncertainty estimation. CCT \cite{ouali2020semi} promotes consistency in feature representations by aligning the predictions of the primary decoder and multiple auxiliary decoders subjected to diverse perturbations.  However, these methods are susceptible to confirmation biases \cite{yang2022st++,arazo2020pseudo} due to the historical update method of EMA \cite{tarvainen2017mean} and the structural sharing among subnets \cite{luo2021efficient}. Additionally, the effectiveness of uncertainty estimation methods based on prediction variance is closely associated with the quantity of predictions generated by various predictors, and they encounter challenges regarding accuracy in estimation and time consumption.
The correction of the erroneous predictions under limited annotation conditions  \cite{wang2023mcf} is also challenging.  
To address this limitation, we propose triple segmentation branches that incorporate evidence learning with different loss functions to enhance prediction diversity. Additionally, we introduce a bidirectional evidence-based cross-supervised learning method to alleviate confirmation bias by reliable knowledge exchange of different branches as shown in Fig. \ref{motivation}. 

From another perspective, pseudo-label-based methods harness the pseudo labels assigned to unlabeled data to enhance the performance of semi-supervised learning. 
Self-training methods \cite{yang2022st++} typically involve pre-training the segmentation model using labeled data and generating pseudo labels for unlabeled data to facilitate joint optimization.
However, relying solely on a single model's predictions for supervision can lead to overfitting and the amplification of inherent biases.
To address this concern, cross-supervised methods typically leverage complementary information between different branches to enhance the effectiveness of semi-supervised learning.
Chen et al. \cite{chen2021semi} propose Cross Pseudo Supervision (CPS), where two differently initialized networks mutually learn from each other’s pseudo labels generated via an argmax operation.  However, this method cannot guarantee that the supervised information from different branches is conducive to the extraction and aggregation of knowledge in semi-supervised training.
To address this issue, we design an evidential conservative branch (ECB) and an evidential progressive branch (EPB) to maintain the disagreement of pseudo labels and the useful guidance of different branches.
ECB aims to generate cautious and conservative segmentation prediction results, with fewer false positive regions. While, the introduction of EFB is to produce progressive and complete prediction results, with the aim of minimizing false negative regions as much as possible, realizing complementary predictions with ECB. 
MC-Net+ \cite{wu2022mutual} introduces three decoders with distinct upsampling strategies to produce slightly varied outputs. The probability predictions of each decoder are sharpened and utilized as pseudo labels to guide the other decoders.
CoraNet \cite{shi2021inconsistency} implements a separate training approach with a conservative branch and a radical branch, specifically addressing certain segmentation regions and uncertain ones using different learning ways. 
However, these methods haven't sufficiently tackled the issue of noise components in pseudo labels. Direct training with pseudo labels tends to make the network overly confident in erroneous predictions.
Some uncertainty-aware semi-supervised segmentation methods excludes the low-quality predictions in pseudo-labels by using uncertainty evaluation methods, such as, Monte-Carlo dropout \cite{yu2019uncertainty}, prediction variance calculation \cite{wu2022mutual}. The multiple sampling operations of Monte-Carlo dropout and multiple predictions from different predictors suffer from estimation accuracy and time-consuming.
Compared with the existing studies, we incorporate the evidence-based uncertainty estimation method via a singular inference process in cross-supervised training by developing evidential conservative branch and evidential progressive branch. Furthermore, we employ evidential fusion to combine decisions from ECB and EPB, supervising the third segmentation branch, evidential fusion branch (EFB). This reduces noise components in the fused pseudo labels, thereby enhancing the efficiency of pseudo-label guided semi-supervised learning.

In summarize, to handle the confirmation bias and error accumulation issues in semi-supervised medical image segmentation guided by pseudo labels, this paper proposes an evidence-based Tri-Branch consistency learning method. We integrate evidence-based optimization, uncertainty guidance, and evidential fusion into the ETC-Net to explore valuable information from unlabeled data. The contributions of this work can be summarized as follows:

\begin{figure}[!t]
	\centerline{\includegraphics[width=3.5in]{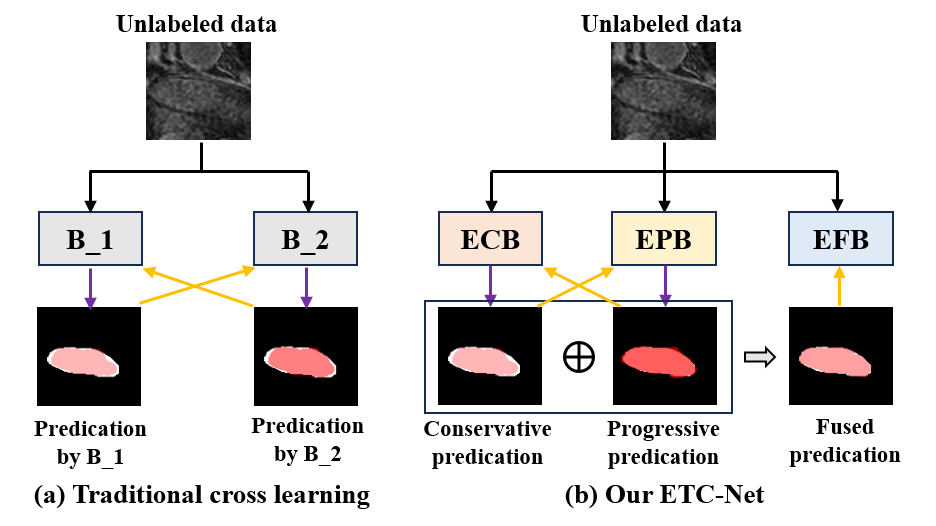}}
	\caption{(a) The traditional cross-supervised learning way. (b) Our proposed ETC-Net. The traditional method is prone to ineffective cross-supervised training when there is no difference between two predictions. Our method can alleviate this issue by generating complementary and reliable ~cross-supervision. Note that the white color in predictions represents ground truth.}
	\label{motivation}
\end{figure}

• Two evidential branches, namely evidential conservative branch (ECB) and evidential progressive  branch (EPB), are proposed for effective semi-supervised training under a low annotation condition, which facilitates the preservation of prediction disagreement in semi-supervised consistency learning. 

• A bi-directional uncertainty-guided cross-supervised training method between ECB and EPB is introduced to enable the reliable exchange of complementary knowledge in unlabeled data.

• An evidential fusion branch, namely EFB, is introduced based on the Dempster-Shafer theory, which is supervised by the fusion results of the belief masses and uncertainty masses generated by ECB and EPB.
Benefiting from the complementary attributes of ECB and EPB and the evidence-based decision fusion strategy, the exploration and transfer of valuable unlabeled knowledge for segmentation improvement is further guaranteed.

• Extensive experiments on three benchmarks attest to the effectiveness of the proposed ETC-Net on semi-supervised medical image segmentation tasks. ETC-Net outperforms other competitive semi-supervised segmentation methods on these benchmarks and significantly improves the segmentation performance compared to the supervised training only using labeled data.

The rest of the paper is organized as follows: In Section \ref{section2}, we briefly introduce the related works. In Section \ref{section3}, we elaborate on the details of the proposed ETC-Net framework. In Section \ref{section4}, we report and analyze the experiment results on three medical image datasets. In Section \ref{Section5}, we conclude this work.

\section{Related work}
\label{section2}
\subsection{Semi-supervised Segmentation}
The objective of semi-supervised segmentation is to improve the segmentation performance by utilizing both labeled and unlabeled data,  which has been highly focused in the medical domain by reducing the requirement for large amounts of annotated data. Pseudo-labeling, adversarial learning, co-training, and consistency regularization are commonly used approaches in semi-supervised segmentation. 
Pseudo-labeling \cite{yang2022st++,li2020self,he2021re,zou2020pseudoseg} methods alternate between generating pseudo labels and updating model parameters. However, the unreliable pseudo-labels lead to performance degradation. Adversarial learning \cite{mittal2019semi,li2020shape} is another popular technique to exploit the unlabeled data by fooling the discriminator into classifying unlabeled data as labeled data. For example, SASSNet \cite{li2020shape} 
utilizes a discriminator to enforce consistent signed distance fields of labeled and unlabeled data.
Recently, there has been a large number of research focusing on designing consistency regularization techniques to leverage unlabeled data, which penalizes the model for generating different outputs for the same input.
The mean-teacher network \cite{tarvainen2017mean,yu2019uncertainty} ensures consistency in both the data and model levels by training a student model and updating a weighted-averaged teacher model, further enforcing the consistent outputs on data subjected to various perturbations between the student and teacher model. 
The integration of uncertainty estimation \cite{yu2019uncertainty,wang2020double} into the mean-teacher network is beneficial to enhancing the reliability of knowledge extraction from unlabeled data, which leads to more precise segmentation supervision signals. However, the sampling cost or resource cost is usually high in uncertainty estimation due to multiple forward times in Monte-Carlo Dropout or multiple model ensembles.
Li, et al \cite{li2020transformation} introduce a transformation-based consistency regularization method which leverages different data augmentations to regularize the network optimization. Lai, et al \cite{lai2021semi} introduce context-aware consistency for semi-supervised semantic segmentation, aiming to address the issue of excessive context usage and improve self-awareness across different contexts.
Pseudo-label based mutual learning methods \cite{wu2022mutual,chen2021semi,fan2022ucc} attempt to make multiple individual learners learn from each other. 
Fan, et al. \cite{fan2022ucc} propose
a cross-head co-training learning framework which employs weak and strong augmentation and takes the merit of co-training consistency constraint. In this paper, we integrate deep evidence learning into the Tri-branch consistency learning network. We devise evidential conservative learning, evidential progressive learning, and evidential decision fusion method to ensure the prediction diversity and boost the reliability of pseudo-labels throughout the consistency learning process.

\subsection{Evidential learning in Image Segmentation}

Recently, a new method for uncertainty estimation, known as Evidential Deep Learning (EDL), has been gaining increasing attention \cite{li2022trustworthy}. The core to EDL is the notion of evidence, which signifies the level of belief or support for different hypotheses within the model's predictions, rooted in Dempster-Shafer theory \cite{dempster1968generalization}. As the model undergoes training, it learns to assign higher evidence to correct predictions and lower evidence to incorrect ones, effectively enhancing the model's performance.
In addition, the evidence from multiple sources can be incorporated to form a belief mass distribution about a particular hypothesis or event.
Benefiting from its advantages, EDL can handle uncertain or incomplete information and is particularly useful when dealing with conflicting or ambiguous evidence \cite{denoeux2000neural}. Many researchers have applied EDL in medical image segmentation tasks for embracing imperfect dataset \cite{huang2021belief,huang2023semi} and fusing multi-modal information \cite{huang2022evidence,diao2021efnet}. 


In addition, many researchers have also engaged in active explorations on using EDL to cope with uncertain, imprecise information sources in various image segmentation tasks, where the uncertainty can be estimated in  a sampling-free way with a reduction of resource cost compared to both Monte-Carlo Dropout \cite{kendall2017uncertainties} and ensemble approaches \cite{xie2016melanoma}.
Tong, et al. \cite{tong2021evidential} integrate the evidential neural network into the fully convolutional network for semantic segmentation.
For the multi-modality segmentation task, Huang, et al. \cite{huang2022evidence} propose a multi-modality evidence fusion method for medical image segmentation which computes a belief function at each voxel for each modality, and merges the discounted evidence of each modality using Dempster’s rule.
EFNet \cite{diao2021efnet} generates the final segmentation results by fusing uncertain evidence from PET and CT. In \cite{sensoy2018evidential}, the belief mass distribution is assigned to a Dirichlet distribution, where a network is trained to predict the hyper-parameters of the Dirichlet distribution and an overall uncertainty mass is quantified using the Dirichlet strength.
TBraTS \cite{zou2022tbrats} applies the evidence-based cross-entropy loss function for trusted medical image segmentation, where the Dirichlet distribution is associated with the class-wise probability and uncertainty of different voxels.
Li, et al \cite{li2022region} extend this idea and propose an evidential soft Dice loss under the Dirichlet prior distribution.  
In this paper, we apply EDL to obtain predictions together with uncertainty quantifications and fuse different predictions from sub-branches, providing a powerful way of combining evidence from multiple predictions to generate more accurate and reliable pseudo-labels under the semi-supervised setting.
Further, we propose a trustworthy evidential triple-branch consistency learning framework based on the evidence theory with Dirichlet distribution for semi-supervised medical image segmentation. We design an evidential conservative branch, an evidential progressive branch, and an evidential fusion branch for reliable and effective semi-supervised medical image segmentation.

\section{Methodology}
\label{section3}

\begin{figure*}[!t]
	\centerline{\includegraphics[width=5.7in]{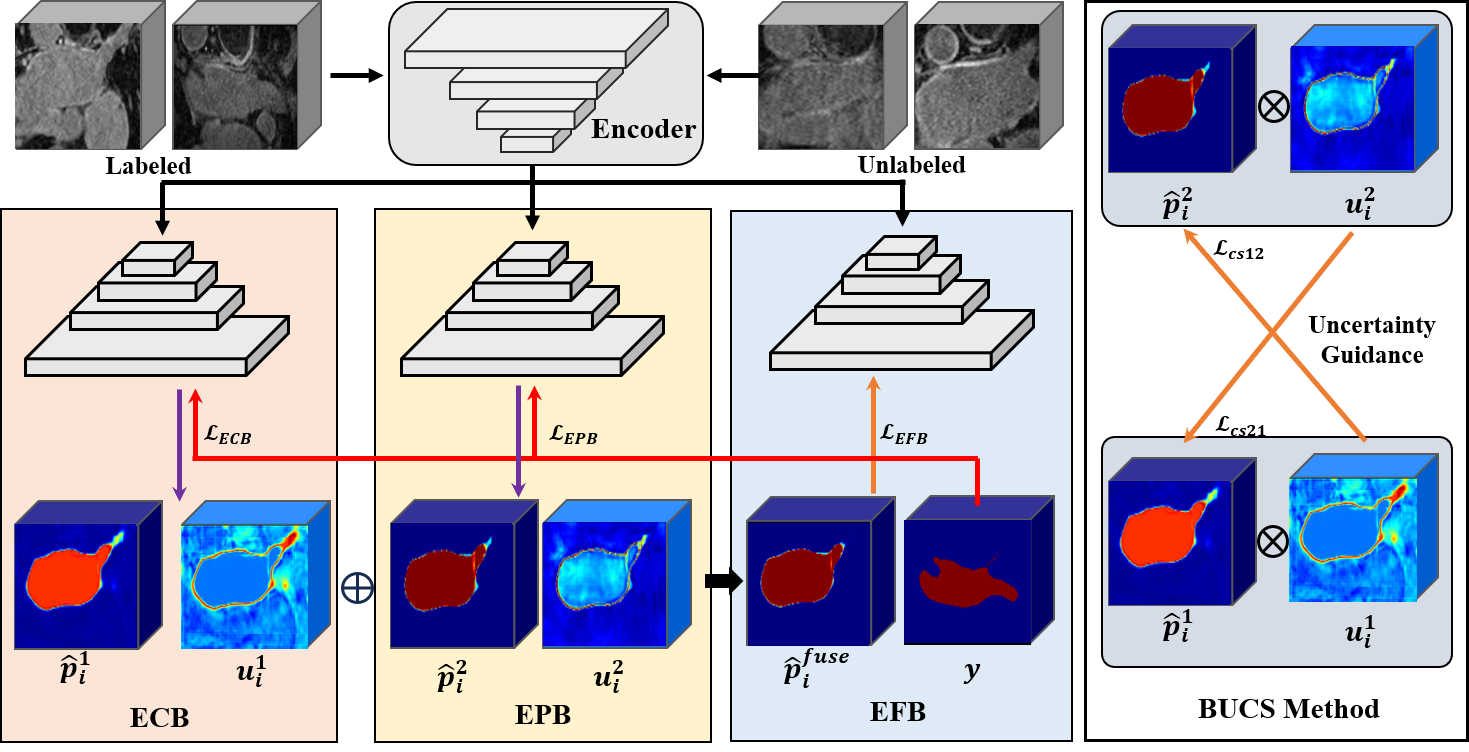}}
	\captionsetup{font=scriptsize}
	\caption{The flowchart of the proposed ETC-Net. $\otimes$ means pixel-wise weighting by uncertain maps $u_i$. $\oplus$ denotes the fusion operation.}
	\label{overallflowchart}
\end{figure*}

The objective of our task is to boost the segmentation performance with the exploitation of labeled data and unlabeled data. The training set is divided into the labeled set $\mathcal{D}_l$ with $N_l$ image-label pairs $(x_l,y_l) \in \mathcal{D}_l$ and the unlabeled set $\mathcal{D}_u$ with $N_u$ images $x_u$,  $N_u \gg N_l$.
In this paper, we propose an evidential tri-branch consistency learning framework for semi-supervised medical image segmentation, which contains a shared encoder and three different sub-branches as shown in Fig. \ref{overallflowchart}.  Specifically, we design two evidential branches with different predictive characteristics, including an evidential conservative branch, which places more emphasis on pixel-level classification accuracy  and an evidential progressive branch emphasizing the overall integrity and completeness of the segmented regions.  
In addition, the evidential learning method can obtain the prediction uncertainty to guide the weighting process in cross-supervised learning. We further introduce another branch, namely evidential fusion branch, which incorporates the knowledge from ECB and EPB by designing an evidential fusion method to generate the pseudo supervision for EFB. 

\subsection{Evidential Conservative Branch}
In ECB, we replace the last softmax function with a nonlinear softplus function. The output evidence vector $\boldsymbol{e_{i}}=\left\{e_{i1},e_{i2},\cdots,e_{iK}\right\}$ is used to assign a Dirichlet distribution with the parameters $\boldsymbol{\alpha_{i}}$ as follows:
\begin{equation}
\alpha_{ik}=e_{ik}+1
\end{equation}
where $K$ is the number of semantic classes, $i$ denotes the pixel index. The Dirichlet strength $S_i$ is defined as:
\begin{equation}
S_i = \sum_{k=1}^{K}\alpha_{ik}
\end{equation}
The predicted class probability vector $\boldsymbol{\hat{p}_{i}}$ can be calculated as: 
\begin{equation}
\hat{p}_{ik} =  \frac{\alpha_{ik}}{S_i}
\label{eq6}
\end{equation}
According to the $K$ outputs of ECB, we formulate the Eq. \ref{uik} with the belief mass $b_{ik}\geq0$ of the $k$-th class for $k=1,\cdots,K$ and the uncertainty mass of the $K$ outputs $u_{i}\geq 0$:
\begin{equation}
\label{uik}
u_{i} + \sum_{k=1}^{K}b_{ik}=1
\end{equation}
where the mass $\left\{b_{i1},b_{i2},\cdots,b_{iK},u_i\right\}$ are formulated as follows:
\begin{equation}
b_{ik} = \frac{e_{ik}}{S_i},u_{i}=\frac{K}{S_i}
\end{equation}
The original cross-entropy loss treats segmentation as a pixel-wise classification problem, 
which has stable gradient update, fast back-propagation speed, and easy convergence.
\begin{equation}
\mathcal{L}_{ce}=\sum_{k=1}^K-y_{ik} \log \left(\hat{p}_{ik}^{1}\right)
\end{equation}
where ``1" in $\hat{p}_{ik}^{1}$  represents that the probability generated from ECB. 
However, the vanilla cross-entropy loss function is sensitive to noise \cite{zhang2018generalized} with SoftMax probability , which cannot provide uncertainty estimation for pseudo labels of unlabeled data. The evidential predictions, conceived from the DST, can provide predictions with uncertainty estimations, which has potential to remove the low-quality pseudo labels by an uncertainty-guided filtering proces. Therefore, we apply the evidence-based cross-entropy loss $\mathcal{L}_{ece}$ in ECB as:
\begin{equation}
\begin{aligned}
\mathcal{L}_{ece} & =\int\left[-\sum_{k=1}^K y_{i k} \log \left(\hat{p}_{ik}^{1}\right)\right] \frac{1}{\mathcal{B}\left(\boldsymbol{\alpha}_i^1\right)} \prod_{k=1}^K \hat{p}_{i k}^{1}{ }^{\alpha_{ik}^{1}-1} d \boldsymbol{\hat{p}_{i}^1} \\
& =\sum_{k=1}^K y_{i k}\left(\psi\left(S_{i}^{1}\right)-\psi\left(\alpha_{i k}^{1}\right)\right),
\end{aligned}
\end{equation}
where $\mathcal{B}\left( \cdot \right)$ represents the multinomial beta function.
$\psi\left( \cdot \right)$ denotes the digamma function.
To ensure that incorrect evidence is reduced to 0, a KL divergence loss function, denoted as $\mathcal{L}_{kl}$, is employed within the context of $\mathcal{L}_{ECB}$:
\begin{equation}
\mathcal{L}_{ECB}=\mathcal{L}_{ece}+ \lambda_{kl}\cdot\mathcal{L}_{kl}
\end{equation}
where $\lambda_{kl}=\min\left(1,t/200\right)$ for stable learning. $t$ denotes the current training iteration. $\mathcal{L}_{kl}$ is defined as follows:
\begin{equation}
\begin{aligned}
\mathcal{L}_{kl}&=K L\left(D\left(\boldsymbol{\hat{p}_{i}^1} \mid \boldsymbol{\tilde{\alpha}_{i}^{1}}\right)\right.  \left.\| D\left(\boldsymbol{\hat{p}_{i}^1} \mid \mathbf{1}\right)\right)\\&=\log \left(\frac{\Gamma\left(\sum_{k=1}^K \widetilde{\alpha}_{i k}^{1}\right)}{\Gamma(K) \prod_{k=1}^K \Gamma\left(\widetilde{\alpha}_{i k}^{1}\right)}\right) \\
&+  \sum_{k=1}^K\left(\widetilde{\alpha}_{ik}^1-1\right)\left(\psi\left(\widetilde{\alpha}_{i k}^1\right)-\psi\left(\sum_{j=1}^K \widetilde{\alpha}_{i j}^1\right)\right)
\end{aligned}
\end{equation}
where $D\left(\boldsymbol{\hat{p}_{i}^{1}} \mid \mathbf{1}\right)$ represents the uniform Dirichlet distribution. $\Gamma\left( \cdot \right)$ denotes the gamma function and $\widetilde{\alpha}_{i k}^1 = y_{ik}+\left({1}-{y}_{ik}\right) \bigodot \alpha_{ik}^{1}$ represents the altered Dirichlet parameters after removal of the true evidence. $\bigodot$ means the element-wise Hadamard product. 
For another, cross-entropy loss may suffer from unsatisfactory performance in dealing with class-imbalanced segmentation. If the number of voxels in a certain category is small, the network may make a conservative decision and ignore this category because its contribution to the loss function is small. Therefore, we further propose a region-based evidential progressive branch, namely EPB, for relatively radical predictions, in the next subsection.
\begin{figure}[!t]
	\centering
	\includegraphics[width=3.45in]{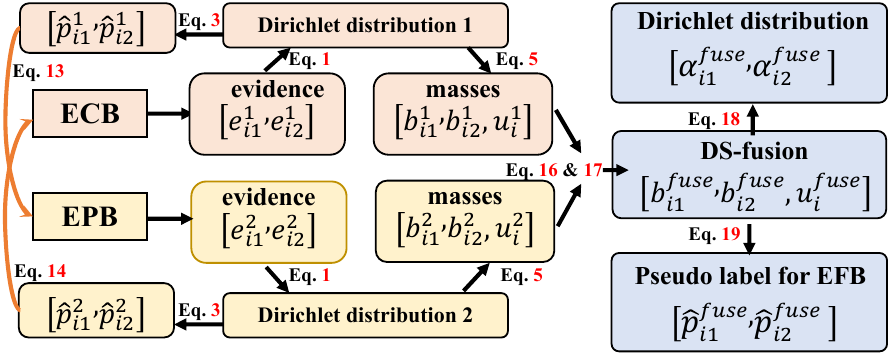}
	\caption{The calculation process of ETC-Net. The orange arrows represent the direction of cross-supervised training.}
	\label{calculationprocess}
\end{figure}
\subsection{Evidential Progressive Branch}
We propose an evidential progressive branch to generate relatively radical predictions by designing an evidential Dice loss.
The Dice loss is based on the Dice Similarity Coefficient (DSC) which calculates the overall similarity between the predicted results and the true labels, defined as:
\begin{equation}
\mathcal{L}_{Dice}=\frac{1}{K}\sum_{k=1}^K\left(1-2 \frac{\sum_{i=1}^{N} \hat{p}_{ik} y_{ik }}{\sum_{i=1}^{N} \hat{p}_{ik}+y_{ik}}\right)
\end{equation}
The difference in the voxel number of different classes is taken into account in $\mathcal{L}_{Dice}$. So correctly predicting pixels in smaller categories will have a greater impact on $\mathcal{L}_{Dice}$, thereby encouraging the model to make more progressive predictions for smaller categories. Thus, the Dice loss is more effective for class-imbalanced problem in segmentation tasks than cross-entropy loss. 
$\mathcal{L}_{Dice}$ can form a complementary advantage with $\mathcal{L}_{ce}$ to ensure the prediction diversity for subsequent cross-supervised learning between different sub-branches.
To endow EFB with the ability to estimate uncertainty, we also assign the evidence output vector $\boldsymbol{e_{i}^2}$ of EPB to the parameters $\boldsymbol{\alpha_{i}^2}$ with a Dirichlet distribution. Then, we propose an evidence-based Dice loss to replace $\boldsymbol{\hat{p}_{i}^2}$ with $\boldsymbol{\alpha_{i}^2}/S_i^2$ in Eq. \ref{eDice}. 
\begin{equation}
\label{eDice}
\mathcal{L}_{EPB}=\frac{1}{K}\sum_{k=1}^K\left(1-2 \frac{\sum_{i=1}^{N} \frac{\alpha_{ik}^2}{S_{i}^2} y_{ik }}{\sum_{i=1}^{N} \frac{\alpha_{ik}^2}{S_{i}^2}+y_{ik}}\right)
\end{equation}

\subsection{Bidirectional Uncertainty-guided Cross Supervision}
Benefiting from the proposed two different evidential branches, ECB and EPB, we design a bidirectional uncertain-guided cross-supervised (BUCS) training method to exchange mutually complementary useful information to alleviate confirmation bias. In BUCS, one branch provides pseudo-labels for the other branch. Due to the deployment of different segmentation loss functions for different branches, prediction diversity can be guaranteed in cross-supervised training procedures. 
Additionally, the pseudo labels are weighted according to uncertainty estimation $u_i$ as Eq. \ref{wi} in evidence learning under a voxel-wise manner. The pseudo-labels with higher uncertainty tend to have small weights in cross-supervised training, which alleviates the negative impact of unreliable parts in the pseudo-labels on model training.
\begin{equation}
\label{wi}
w_i^1 = 1-u_i^1, w_i^2 = 1-u_i^2
\end{equation}
We select $\mathcal{L}_2$ as the optimization constraint.
Therefore, the cross-supervised loss from ECB to EPB which uses the probability prediction $p_i^1$ of ECB to supervise $p_i^2$ of EPB, which is formulated as: 
\begin{equation}
\mathcal{L}_{cs12} = \frac{1}{K}\sum_{k=1}^{K} w_i^1 \cdot \left\|\hat{p}_{ik}^1-\hat{p}_{ik}^2\right\|_2
\label{eq16}
\end{equation}
Similarly, the cross-supervised loss from EPB to ECB is defined as:
\begin{equation}
\mathcal{L}_{cs21} = \frac{1}{K}\sum_{k=1}^{K} w_i^2 \cdot \left\|\hat{p}_{ik}^2-\hat{p}_{ik}^1\right\|_2
\label{eq17}
\end{equation}
Note that the cross-supervised loss is used to train both labeled data and unlabeled data.
Lastly, the bidirectional cross-supervised loss $\mathcal{L}_{BUCS}$ is written as:
\begin{equation}
\mathcal{L}_{BUCS}= \mathcal{L}_{cs12} + \mathcal{L}_{cs21}
\end{equation}
\subsection{Evidential Fusion Branch}
To further reduce the error predictions in pseudo labels and mine useful information in the unlabeled data, we introduce an evidential fusion branch by exploiting the complementary characteristic of ECB and EPB. 
$p_k^3$ denotes the segmentation probability prediction of EFB.
Specifically, we generate the pseudo supervision for $p_k^3$ in a cross-supervised manner using Dempster-Shafer theory following Eq. \ref{eqbik} to Eq. \ref{pfuse}, which fuse the predictions of ECB and EPB with three steps. Firstly, the fusion of belief masses and uncertainty masses are inferred according to Eq. \ref{eqbik} and Eq. \ref{ufuse}: 
\begin{equation}
b_{ik}^{fuse}=b_{ik}^{1 \oplus 2}=\frac{1}{1-Q}\left(b_{ik}^1 b_{ik}^2+b_{ik}^1 u_i^2+b_{ik}^2 u_i^1\right)
\label{eqbik}
\end{equation}
\begin{equation}
\label{ufuse}
u_i^{fuse}=u_i^{1 \oplus 2}=\frac{1}{1-Q} u_i^1 u_i^2
\end{equation}
where $Q=\sum_{k\neq j} b_{ik}^1 b_{ij}^2$ represents the conflict degree of two belief masses of ECB and EPB.

Secondly, the parameters of the fused Dirichlet distribution are calculated by Eq. \ref{alphafuse}. 
\begin{equation}
\label{alphafuse}
e_{ik}^{fuse}=\frac{K b_{ik}^{fuse}}{u_i^{fuse}}, \alpha_{ik}^{fuse}=e_{ik}^{fuse}+1 .
\end{equation}

Lastly, the pseudo supervision for EFB is defined as follows:  
\begin{equation}
\label{pfuse}
\hat{p}_{ik}^{fuse}=\frac{\alpha_{ik}^{fuse}}{S_i^{fuse}}=  \frac{\alpha_{ik}^{fuse}}{\sum_{k}\alpha_{ik}^{fuse}}
\end{equation}
Therefore, the loss constraint of EFB is formulated as:
\begin{equation}
\mathcal{L}_{EFB}= \sum_{k=1}^{K}  \left\|\hat{p}_{ik}^{fuse}-\hat{p}_{ik}^3\right\|_2
\end{equation}
\subsection{Training Objective of ETC-Net}
To summarize, the total loss function of ETC-Net compromises four parts, the supervised loss $\mathcal{L}_{ECB}$ and $\mathcal{L}_{EPB}$,  the bidirectional cross-supervised loss $\mathcal{L}_{BUCS}$ for labeled and unlabeled data, and the loss constraint for EFB $\mathcal{L}_{EFB}$ for labeled and unlabeled data.
\begin{equation}
\mathcal{L}_{ETC} = \mathcal{L}_{ECB} + \mathcal{L}_{EPB}+ \lambda \left( \mathcal{L}_{BUCS} + \mathcal{L}_{EFB}\right)  
\end{equation}
where $\lambda$ is a time-dependent weighting parameter with Gaussian ramp-up curve \cite{tarvainen2017mean}, which balances the evidential supervised loss $\mathcal{L}_{ECB}$ and $\mathcal{L}_{EPB}$ and the cross-supervised loss constraint, $\mathcal{L}_{BUCS}$ and $\mathcal{L}_{EFB}$.
\begin{equation}
\label{eqlamda}
\lambda=w_{max} \cdot e^{\left(-5\left(1-t / t_{\max }\right)^{2}\right)}
\end{equation}
$w_{max}$ in Eq. \ref{eqlamda} is empirically set to 0.1. $t$ is the current number of iteration and $t_{max} = 1000$ is the maximum number of iteration.

\section{Experiments}
\label{section4}
\subsection{Datasets}
We validate the effectiveness of our method on three public benchmarks, including LA dataset \cite{xiong2021global}, Pancreas-CT dataset \cite{clark2013cancer}, and ACDC dataset \cite{bernard2018deep}. 
LA dataset is the benchmark in the 2018 atrial segmentation challenge, which comprises 100 gadolinium-enhanced MRI scans with an isotropic resolution of $0.625 \times 0.625 \times 0.625mm^3$. 
We employ a split of 80 training scans and 20 test scans following \cite{yu2019uncertainty}. All scans are center-cropped surrounding the tissue region.
The NIH Pancreas-CT dataset includes 82 abdominal CT scans, and the resolution of axial slices is $512 \times 512$. We re-sample the data to have an isotropic resolution of $1 \times 1 \times1mm^3$.
Following the previous work \cite{luo2021semi,wu2022mutual},
we employ a division of 62 training scans and 20 test scans. For pre-processing, we crop 3D samples from the original data to preserve the target region and normalize it to zero mean and unit variance. 
The Automated Cardiac Diagnosis Challenge (ACDC) dataset contains 200 annotated short-axis cardiac cine-MRI scans from 100 subjects. All scans are randomly divided into 140 training scans, 20 validation scans, and 40 test scans following the previous work \cite{wu2022mutual}. The 3D scans in ACDC dataset are converted into 2D slices. 
Unlike the 3D segmentation task on LA and Pancreas-CT datasets, we extend ECT-Net to a 2D version to segment three tissues, i.e.  myocardium, left ventricle and right ventricle (Myo, RV, LV) from these 2D slices.
In this study, we compare other popular methods on LA, Pancreas-CT and ACDC datasets with a 10\% labeled ratio. 

\subsection{Implementation Details and Evaluation Metrics} 
In 3D segmentation training, we randomly crop 3D patches with the size of $112 \times 112 \times80$ on LA dataset and $96 \times 96 \times96$ on Pancreas-CT dataset, respectively. The intensities of the cropped 3D patches are normalized to zero mean and unit variance.
The batch size is set to 4, including 2 labeled patches and 2 unlabeled patches. V-Net is employed as the backbone for 3D segmentation. 
For 2D segmentation training, each slice is resized to $256 \times 256$ and normalized as zero mean and unit variance. U-Net is employed as the backbone for 2D segmentation. 
For all segmentation tasks, we employ an SGD optimizer with an initial learning rate of 0.1 and use a poly learning rate decay policy.
The number of training iterations is 30000.
In the testing stage, we adopt a sliding window strategy with the fixed step size of $18 \times18 \times 4$ and $16 \times 16 \times 16$ on LA dataset and Pancreas-CT dataset, respectively. The patch predictions are re-composed to form the entire results.
We use four evaluation metrics to evaluate the performance, including DSC, JAC, the average surface distance (ASD), and the 95\% Hausdorff Distance (95HD).
All experiments are implemented by PyTorch with four NVIDIA GeForce RTX 3090 Ti GPUs.

\begin{table}[h]
	\caption{Comparison with other methods on LA dataset. The red font denotes the best performance. The blue font denotes the second-best performance.}\label{LAres}
	\centering
	\resizebox{\linewidth}{!}{
		\begin{tabular}{@{}lcccccc@{}}
			\toprule
			\multirow{2}{*}{Method} & \multicolumn{2}{c}{Scans used} & \multicolumn{4}{c}{Metrics}                            \\ \cmidrule(r){2-3} \cmidrule(r){4-7} 
			& Labeled      & Unlabeled      & DSC(\%)$\uparrow$ & JAC(\%)$\uparrow$ & ASD(voxel)$\downarrow$  & 95HD(voxel)$\downarrow$        \\ \midrule
			V-Net                   & 8 (10\%)          & 0              & 86.03    & 76.08        & 2.45       & 10.55             \\
			V-Net                   & 80 (100\%)         & 0              & 92.27     & 85.69        & 1.30        & 5.25                    \\ \midrule
			MT \cite{tarvainen2017mean}                     & 8           &72             & 87.09     & 77.44        & 2.87        & 10.27        \\
			UA-MT \cite{yu2019uncertainty}                  &8           &72             & \textcolor{blue}{88.43}    & \textcolor{blue}{79.40}        & 3.35        &12.44                \\
			SASSNet \cite{li2020shape}                 & 8           &72             &85.99           &75.71              &\textcolor{blue}{2.00}             &13.10                           \\
			DTC \cite{luo2021semi}                   & 8           & 72             & 87.36     & 77.75       & 2.02        & 9.77                   \\
			URPC \cite{luo2021efficient}                    & 8           & 72             &87.05     & 77.33        &2.12             &9.91                     \\
			MC-Net+ \cite{wu2022mutual}                 & 8           & 72             & 86.34          &76.27              &2.12             &9.92                            \\
			CPS \cite{chen2021semi}&8&72&88.06&77.87&2.13&8.33\\ ASE-Net \cite{9966841}&8&72&86.81&76.82&3.96&15.08\\
			CoraNet \cite{shi2021inconsistency}&8&72&87.97&78.74&2.89&11.09\\MCF\cite{wang2023mcf}&8&72&87.06&77.83&2.67&\textcolor{blue}{7.81} \\
			ETC-Net (Ours)           & 8           & 72             & \textcolor{red}{91.15} & \textcolor{red}{83.80} & \textcolor{red}{1.65} & \textcolor{red}{5.45}                        \\ \bottomrule
	\end{tabular}}
\end{table}
\begin{table}[h]
	\caption{Comparison with other methods on Pancreas-CT dataset.}
	\centering
	\resizebox{\linewidth}{!}{
		\begin{tabular}{@{}lcccccc@{}}
			\toprule
			\multirow{2}{*}{Method} & \multicolumn{2}{c}{Scans used} & \multicolumn{4}{c}{Metrics}                            \\ \cmidrule(r){2-3} \cmidrule(r){4-7} 
			& Labeled      & Unlabeled      & DSC($\%$)$\uparrow$ & JAC(\%)$\uparrow$ & ASD(voxel)$\downarrow$  & 95HD(voxel)$\downarrow$        \\ \midrule
			V-Net                   & 6 (10\%)          & 0              & 68.24    & 53.24        & 2.21       & 13.76             \\
			V-Net                   & 62 (100\%)         & 0              & 82.99     & 71.27        &1.15        &4.67                    \\ \midrule
			MT \cite{tarvainen2017mean}                     & 6           &56              & 73.72     & 59.25       &2.44        &15.18                  \\
			UA-MT \cite{yu2019uncertainty}                  &6           &56      & \textcolor{blue}{75.02}     & \textcolor{blue}{61.14}        & \textcolor{blue}{1.68}        & \textcolor{blue}{13.84}               \\
			SASSNet \cite{li2020shape}                 & 6           &56             &71.76          &57.53              &\textcolor{red}{1.65}             &15.27                           \\
			DTC \cite{luo2021semi}                   & 6           & 56             & 64.71     & 50.05        & 2.20        & 22.70                   \\
			URPC \cite{luo2021efficient}                    & 6           & 56             & 72.05     & 57.85        &3.05             &13.13                    \\
			MC-Net+ \cite{wu2022mutual}                  & 6          & 56             & 71.68          &56.91              &3.62             &19.49                            \\
			CPS \cite{chen2021semi}&6&56&74.86&60.72&1.76&14.24\\ ASE-Net \cite{9966841}&6&56&69.92&54.58&{5.28}&19.74\\
			CoraNet \cite{shi2021inconsistency}&6&56&72.64&57.99&5.13&22.64\\MCF \cite{wang2023mcf}&6&56&69.00&54.80&1.90&14.23\\
			ETC-Net (Ours)           & 6           & 56             & \textcolor{red}{77.15}     & \textcolor{red}{63.54}        & {1.81}        & \textcolor{red}{9.45}  
			\\ \bottomrule
	\end{tabular}}
	\label{Pancreastab}
\end{table}
\subsection{Comparison with Other Semi-supervised Methods}
We compare the proposed framework with 10 popular semi-supervised segmentation methods, fully-supervised training with 10\% labeled data and 100\% labeled data.
\begin{figure*}[!t]
	\centering
	\includegraphics[width=7.1in]{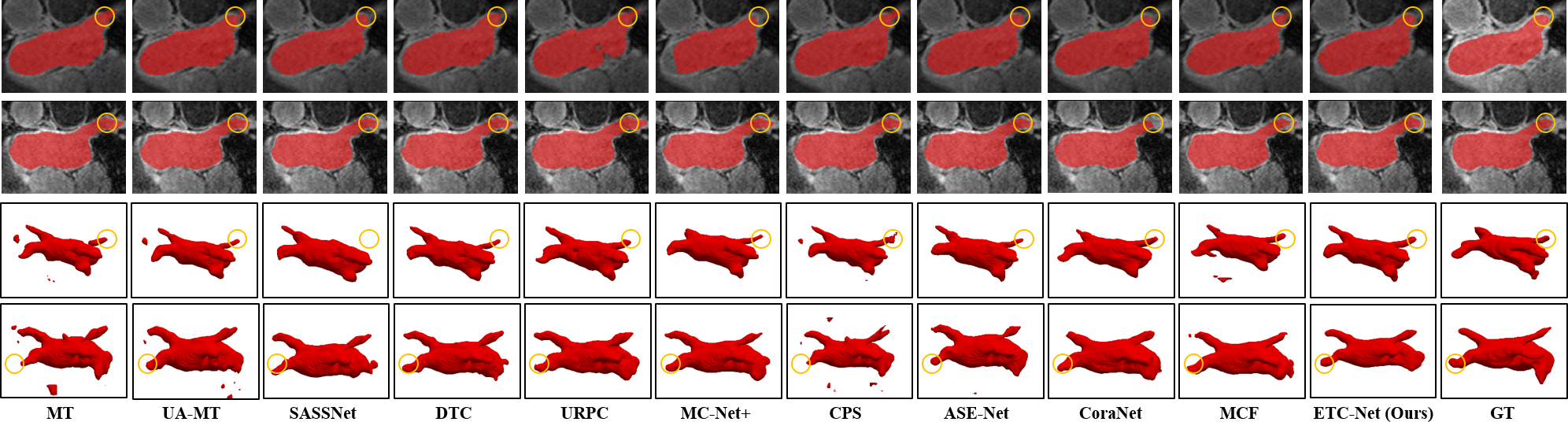}
	\caption{2D and 3D Visualized results of semi-supervised segmentation results on LA dataset for left atrium segmentation.}
	\label{semiLAvis}
\end{figure*} 
\begin{figure*}[!t]
	\centering
	\includegraphics[width=7.1in]{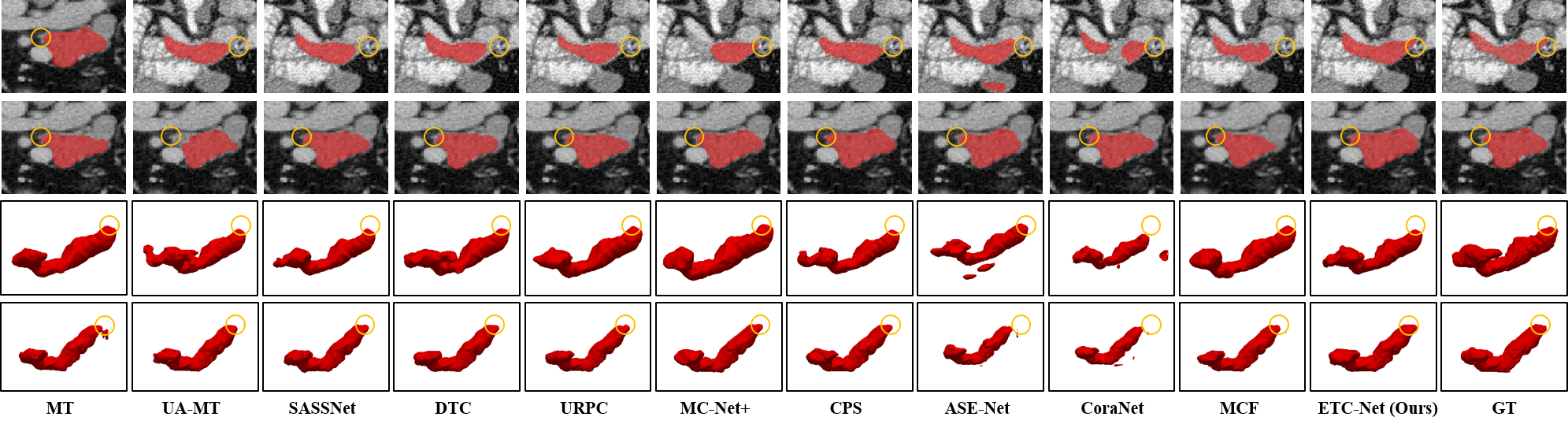}
	\caption{2D and 3D Visualized results of semi-supervised segmentation results for pancreas segmentation.}
	\label{predPancvisual}
\end{figure*}
\begin{figure*}[h]
	\centering
	\includegraphics[width=7.1in]{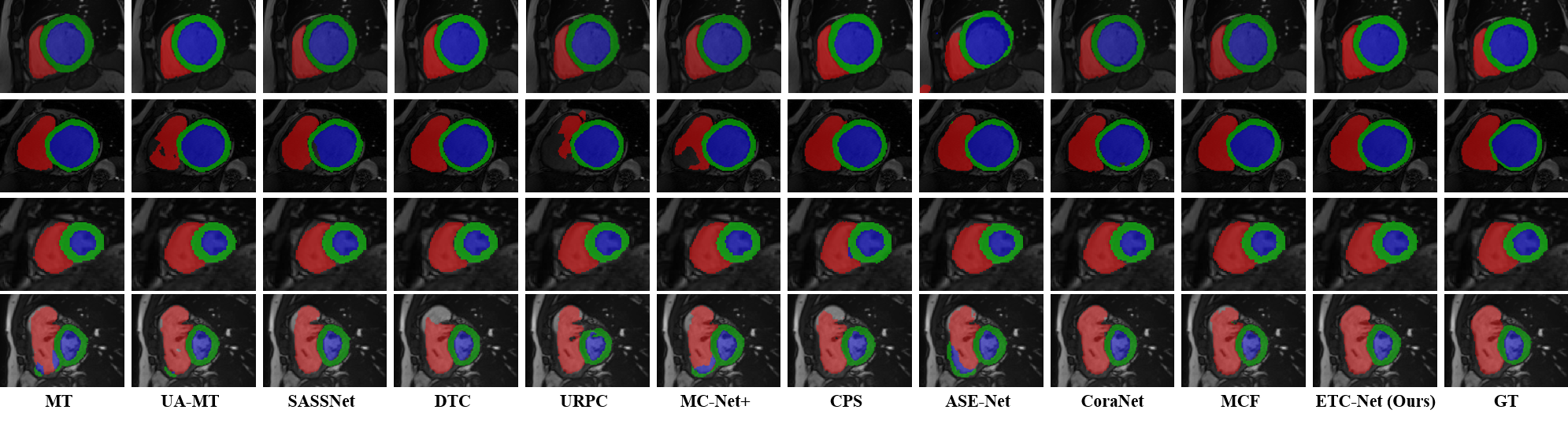}
	\caption{Visualized results of semi-supervised segmentation results for RV (red), Myo (green), and LV (blue).}
	\label{predACDCvisual}
\end{figure*}
\subsubsection{Semi-supervised results on LA dataset}
Based on the experimental findings presented in Table \ref{LAres}, our proposed ETC-Net demonstrates superior performance compared to other methods in semi-supervised left atrium segmentation.
Compared to the baseline method V-Net, which solely relies on labeled data, ETC-Net exhibits significant improvements across all metrics. Notably, ETC-Net achieves an impressive JAC score of 83.80\%, surpassing V-Net's performance of 76.08\% by 7.72\% using 8 labeled samples alone. This demonstrates the efficacy of ETC-Net for extracting valuable unlabeled information by integrating evidence-based optimization, uncertainty guidance, and evidential decision fusion.
In addition, ETC-Net achieves 2.72\% and 4.40\% higher DSC and JAC compared to the second-best UA-MT. Additionally, ETC-Net achieves the lowest ASD and 95HD values, which indicates that ETC-Net achieves improved semi-supervised left atrium segmentation results and signifies enhanced boundary delineation.
We also visualize 2D and 3D left atrium segmentation samples in Fig. \ref{semiLAvis}. ETC-Net generates more consistent boundaries and more complete segmentation structures compared to ground truth.

\subsubsection{Semi-supervised results of pancreas segmentation}
Table \ref{Pancreastab} presents the pancreas segmentation results achieved by our proposed method and competitors on DSC, JAC, ASD, and 95HD.
As shown in Table \ref{Pancreastab},  ETC-Net consistently outperforms all other comparative methods, demonstrating the superiority of our method.
Specifically, compared with the baseline V-Net only using 6 labeled cases, our method achieves significant improvements of 8.91\% and 10.30\% on DSC and JAC,  respectively. Compared with the second-best method UA-MT, our ETC-Net improves DSC, JAC, and 95HD by 2.13\%, 2.40\%, and 4.39, respectively. These results demonstrate that the complementary cross-supervised learning and evidence fusion between ECB and EPB contribute to extracting valuable information from unlabeled data for pancreas segmentation improvement.
Additionally, as depicted in Fig. \ref{predPancvisual}, the visualizations of some pancreas segmentation results are presented in both 2D and 3D formats. The segmentation results achieved by ETC-Net exhibit a high alignment with the ground truths, thus highlighting the effectiveness of ETC-Net in pancreas segmentation. Particularly noteworthy are the accurate segmentation details within the yellow circles in Fig. \ref{predPancvisual}, further validating our method's ability to generate high-quality segmentation masks with well-defined boundaries.
\begin{table*}[!t]
	\caption{Comparision with other methods on the ACDC test set. DSC (\%) and ASSD (mm) are reported with 14 labeled scans and 126 unlabeled scans for semi-supervised training. The bold font represents the best performance.}
	\resizebox{\linewidth}{!}{
		\begin{tabular}{@{}c|c|cccc|cccc|cccc|cccc@{}}
			\hline
			\multirow{2}{*}{Method} & {Scans } & \multicolumn{4}{c|}{RV}                                           & \multicolumn{4}{c|}{Myo}                                         & \multicolumn{4}{c|}{LV}                                          & \multicolumn{4}{c}{Avg} \\ 
			& L/U       & DSC$\uparrow$ & JAC$\uparrow$ & ASD$\downarrow$  & 95HD$\downarrow$                    & DSC$\uparrow$ & JAC$\uparrow$ & ASD$\downarrow$  & 95HD$\downarrow$           & DSC$\uparrow$ & JAC$\uparrow$ & ASD$\downarrow$  & 95HD$\downarrow$                  & DSC$\uparrow$ & JAC$\uparrow$ & ASD$\downarrow$  & 95HD$\downarrow$                \\ \hline
			U-Net                   & 14/0 (10\%)           & 82.63               &71.97                            &2.06                             &7.93                        &78.84                           &65.35                           &3.22                            &13.50       &86.37  &77.44 &4.35 &18.98 &82.61 &71.58 &3.21 &13.47   \\
			U-Net                   & 140 (100\%)           &91.13               &84.29                           &0.58                            &2.19             &89.35&80.92&0.46&1.02&95.03&90.70&0.36&1.49                            &91.83                          &85.31                           &0.47       &1.57      \\ \hline
			MT \cite{tarvainen2017mean}                   & 14/126          &86.09              & 76.87                         &1.11                             &2.73&84.78 &73.82                           &1.64 &6.32                           &90.13                  &83.08 &2.35&8.36                           &87.00      &77.92&1.70&5.80        \\
			UAMT \cite{yu2019uncertainty}              & 14/126            &\textcolor{red}{86.65}              & \textcolor{red}{77.57}                          &0.90                             &3.37&84.98 &74.20                           &1.78 &7.82                           &89.98                  &82.67 &3.78&13.81                           &\textcolor{blue}{87.20}      & \textcolor{blue}{78.15}&2.16&8.34    \\
			SASSNet \cite{ouali2020semi}                     & 14/126        &85.20              &75.34                          &2.38                            &11.58                           &85.06                           &74.38                            &1.37                            &6.66       &90.12 &82.86&1.77&\textcolor{blue}{6.08}&86.79&77.52 &1.84&8.10    \\DTC \cite{luo2021semi}&14/126&85.28&76.08&\textcolor{blue}{0.70}&\textcolor{red}{1.89}&85.33&74.77&1.22&5.79&90.32&83.21&2.51&8.15&86.97&78.02&\textcolor{blue}{1.48}&\textcolor{blue}{5.28} \\
			URPC \cite{luo2021efficient}                  & 14/126            & 80.20             & 70.12                           & {1.18} &4.53                            &84.92                           &74.14 &1.76                           & 8.66                                &90.08 &82.77&3.37&7.22 &85.06&75.45&2.10&6.80    \\
			MC-Net+ \cite{wu2022mutual}                  & 14/126            &84.49             &74.64                           &1.26                             &3.95                            &83.86                           &72.52                           &2.45                           &8.63       &88.82&81.09&3.81&15.65&85.73&76.08&2.51&9.41       \\
			
			CPS \cite{chen2021semi}            & 14/126            &81.31             &70.22                           &1.13             &3.53             &81.83            &69.66            &2.02             &5.85       & 87.19&78.82&3.47&10.66&83.45&72.90&2.21&6.68   \\ASE-Net \cite{9966841}&14/126&84.76&74.53&{3.87}&16.63&83.89&72.70&1.61&6.59&\textcolor{blue}{91.41}&\textcolor{blue}{84.95}&\textcolor{blue}{1.64}&6.14&86.68&77.39&2.37&9.78\\
			CoraNet \cite{shi2021inconsistency}&14/126&83.05&72.72&\textcolor{red}{0.46}&\textcolor{blue}{2.40}&\textcolor{blue}{85.61}&\textcolor{blue}{75.18}&\textcolor{blue}{1.17}&\textcolor{blue}{5.34}&89.86&82.33&3.23&11.42&86.17&76.74&1.62&6.38\\MCF \cite{wang2023mcf}&14/126&83.76&73.16&3.13&10.29&81.58&69.29&1.87&8.40&88.33&80.11&2.14&7.55&84.55&74.18&2.38&8.74\\
			ETC-Net (Ours)                 & 14/126                      & \textcolor{blue}{86.52} & \textcolor{blue}{77.28}                            & \textcolor{black}{1.48} & {5.52} & \textcolor{red}{85.66} & \textcolor{red}{75.26} & \textcolor{red}{0.82}       & \textcolor{red}{3.50}&\textcolor{red}{92.07}&\textcolor{red}{85.68}&\textcolor{red}{1.43}&\textcolor{red}{4.25}&\textcolor{red}{88.08}&\textcolor{red}{79.40}&\textcolor{red}{1.24}&\textcolor{red}{4.42}\\ \hline
	\end{tabular}} \label{ACDCtable}
\end{table*}
\subsubsection{Semi-supervised results on ACDC dataset}
As shown in Table \ref{ACDCtable}, we compare ETC-Net with other competitive semi-supervised segmentation methods for the cardiac structure, RV, Myo, and LV segmentation. Our method performs best on all metrics, achieving 88.08\%, 79.40\%, 1.24 and 4.42 on the average DSC, JAC, ASD and 95HD, respectively. Our method achieves 0.88\% and 1.25\% higher scores than the second-place UA-MT on DSC and JAC, respectively.  These findings indicate that ETC-Net takes advantages of uncertainty estimation in evidential learning and knowledge compensation of different branches to achieve stable and improved semi-supervised segmentation performance on ACDC dataset. Additionally, it outperforms the second-best DTC by 0.24 and 0.86 on ASD and 95HD, respectively, demonstrating that our approach not only delivers accurate cardiac segmentation performance but also maintains a more anatomically consistent segmentation shape with ground truth.
We also provide visualized samples in Fig. \ref{predACDCvisual} to qualitatively illustrate the effectiveness of ETC-Net on ACDC dataset.
Our method showcases its efficacy through aligned boundaries and complete structures compared to ground truth annotation. For example, some methods like CPS, and DTC fail to segment the complete RV structure in the fourth case. The proposed ETC-Net showcases improved segmentation integrity of the RV tissue, validating the effectiveness of cross-supervised training and evidence fusion.
\subsubsection{Segmentation results of different branch}
As presented in Table \ref{threeclassifier}, EFB achieves superior segmentation performance to ECB and EPB with the distillation of the evidence fusion of $\hat{p}_i^1$ and $\hat{p}_i^2$. For example, EFB achieves 0.92\% and 0.98\% improvements on DSC in the LA dataset compared to ECB and EPB, respectively. These improvements indicate the proposed evidence fusion method achieves complementarity between two branches, ultimately provides high-quality supervisory information for EFB.
In addition, the ensembles of triple branches generates higher DSC and JAC on three segmentation tasks than any individual branch, which demonstrates the proposed ETC-Net can effectively exploit and utilize complementary and valuable information from each branch, leading to a synergistic improvement in segmentation quality.
\begin{table}[h]
	\centering
	\caption{Segmentation results of different branches on three datasets. We report the average performance of ACDC dataset.}
	\resizebox{\linewidth}{!}{
		\begin{tabular}{@{}cccccc@{}}
			\toprule
			Dataset                      & Branch   & DSC(\%)$\uparrow$   & JAC(\%)$\uparrow$   & ASD(voxel)$\downarrow$  & 95HD(voxel)$\downarrow$  \\ \midrule
			\multirow{4}{*}{LA}          &ECB  & 89.26 & 80.74 & 1.69 & 8.04  \\
			& EPB  & 89.20 & 80.64 & 1.70 & 7.56  \\
			& EFB  & \textcolor{blue}{90.18} & \textcolor{blue}{82.21} & \textcolor{blue}{1.68} & \textcolor{blue}{7.34}  \\
			& All       & \textcolor{red}{91.15} & \textcolor{red}{83.80} & \textcolor{red}{1.65} & \textcolor{red}{5.45}  \\ \midrule
			\multirow{4}{*}{Pancreas-CT} &ECB  & 74.94 & 60.46 & 2.60 & 11.05 \\
			&EPB  & 75.68 & 61.66 & 2.35 & 10.27 \\
			&EFB  & \textcolor{blue}{76.52} & \textcolor{blue}{62.61} & \textcolor{blue}{1.94} & \textcolor{blue}{9.77}  \\
			& All       & \textcolor{red}{77.15} & \textcolor{red}{63.54} & \textcolor{red}{1.81} & \textcolor{red}{9.45}  \\ \midrule
			& ECB & 86.81      & 77.52      &1.52      &6.47       \\
			& EPB  &87.03       &77.94       &1.72      &7.12       \\
			ACDC                         &EFB  &\textcolor{blue}{87.85}       &\textcolor{blue}{78.58}       &\textcolor{blue}{1.37}      &\textcolor{blue}{5.13}       \\
			& All       &\textcolor{red}{88.08}&\textcolor{red}{79.40}&\textcolor{red}{1.24}&\textcolor{red}{4.42}   \\ \bottomrule
	\end{tabular}}
	\label{threeclassifier}
\end{table}

\begin{figure}[!h]
	\centering
	\includegraphics[width=3in]{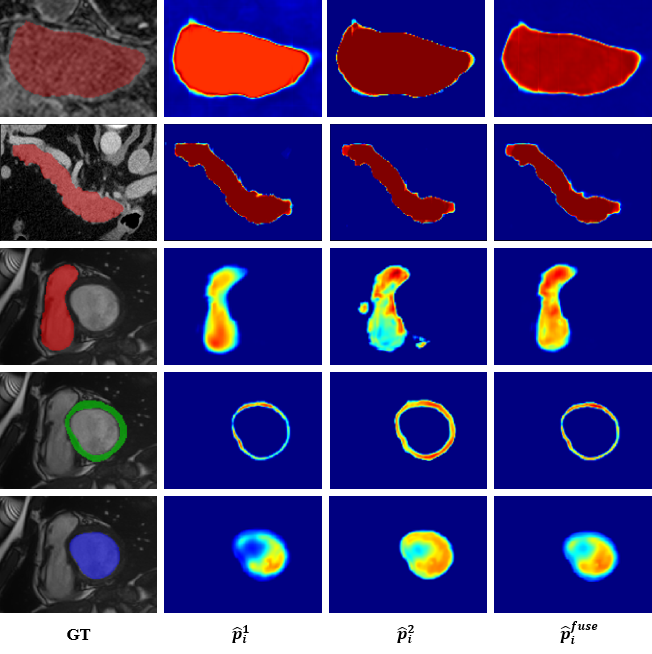}
	\caption{The second column and the third column display the pseudo labels generated by ECB and EPB, $\hat{p}_i^1$ and $\hat{p}_i^2$ in Eq. \ref{eq6}, respectively. The last column shows the fused pseudo label $\hat{p}_i^{fuse}$ for EFB in Eq. \ref{pfuse}.}
	\label{pseudoclassifier}
\end{figure}

We visualize the pseudo labels of bidirectional cross-supervised training in Fig. \ref{pseudoclassifier}, where $\hat{p}_i^1$ is generated by ECB and $\hat{p}_i^2$ is generated by EPB. The ECB tends to produce more precise tissue predictions, however, it may have a lower recall rate, leading to some  true positive regions are not entirely captured. For example, in the fifth row and second column, concerning the pseudo-label of the LV class, there are instances where ECB doesn't entirely succeed in predicting all the true positive regions.
On the other hand, EPB tends to produce more complete foreground segmentation results, but it may introduce some false positive regions. For instance, in the third row and third column, there are scenarios where the pseudo-label of RV class contains anatomically unreasonable regions.
The distinctive characteristics of ECB and EPB foster the diversity between the two branches within the framework of bidirectional cross-supervised learning, thereby ensuring the efficacy of the learning process. Furthermore, the decision fusion based on $\hat{p}_i^1$ and $\hat{p}_i^2$ from both branches enables the advantageous complementarity, furnishing the EFB with a dependable supervisory signal $\hat{p}_i^{fuse}$, which fully leverages beneficial information from unlabeled data for efficient and reliable semi-supervised training.
\begin{table}[!t]
	\centering
	\caption{Ablation study of the different loss components of ETC-Net. We report the left atrium segmentation performance on DSC.}
	\label{ablaloss}
	\resizebox{\linewidth}{!}{
		\begin{tabular}{@{}ccccccccc@{}}
			\toprule
			$\mathcal{L}_{ECB}$&$\mathcal{L}_{EPB}$ &$\mathcal{L}_{cs12}$&$\mathcal{L}_{cs21}$&$\mathcal{L}_{EFB}$                                                                                                                       & ECB & EPB & EFB & All \\ \midrule
			\checkmark&\ding{55}&\ding{55}&\ding{55}&\ding{55}                                               & 86.41              &-               & -              &-        \\
			\ding{55}&\checkmark&\ding{55}&\ding{55}&\ding{55}                                                                                        &-               & 86.29              & -              & -       \\\checkmark&\checkmark&\checkmark&\ding{55}&\ding{55}
			& 88.46              &88.47               & -              &-        \\\checkmark&\checkmark&\ding{55}&\checkmark&\ding{55}
			&88.38               & 88.46              &-               &-        \\\checkmark&\checkmark&\checkmark&\checkmark&\ding{55}
			&\textcolor{blue}{89.10}               &\textcolor{blue}{89.12}               &-               &-        \\\checkmark&\checkmark&\checkmark&\checkmark&\checkmark&\textcolor{red}{89.26}&\textcolor{red}{89.20}&\textcolor{red}{90.18}&\textcolor{red}{91.15} \\\bottomrule
		\end{tabular}}
\end{table}

\subsection{Ablation study}
We conduct ablation studies on LA dataset. The contribution of different loss components and the uncertainty-weighting in the loss functions are studied with 8 labeled cases and 72 unlabeled cases in subsection \ref{Ablcom}.
\subsubsection{Ablation study of BUCS}
\label{Ablcom}
After introducing the supervised loss $\mathcal{L}_{cs12}$ from ECB to EPB, as shown in Table \ref{ablaloss}, the DSC of EPB is improved from 86.29\% to 88.47\%, which demonstrates that ECB can provide valuable and complementary information to EPB.
After introducing the supervised loss $\mathcal{L}_{cs21}$ from EPB to ECB, the DSC of ECB is improved from 86.41\% to 88.38\%, which validates that EPB can also provide beneficial supervision to ECB. However, the pseudo-labels from EPB may have some false positive regions, affecting the segmentation accuracy.

After introducing the bidirectional cross-supervised loss $\mathcal{L}_{cs21}$ and $\mathcal{L}_{cs12}$ in the fifth row, the DSC of ECB increases by 2.69\% compared to the first row in Table \ref{ablaloss}, and the DSC of EPB is improved by 2.83\% compared to the second row, which demonstrates that the proposed BUCS method can leverage the complementary strengths of both branches, mitigating the cognitive biases and the error accumulation problem to provide valuable mutual supervision information from unlabeled data, thereby enhancing semi-supervised segmentation performance.
In addition, the visualization of pseudo-label results in the second column and the third column of Fig. \ref{pseudoclassifier} also supports the above experimental analysis.
\subsubsection{Ablation study of EFB}
EFB distills the valuable and complementary unlabeled knowledge from both ECB and EPB. After the addition of the loss component $\mathcal{L}_{EFB}$ for EFB, the DSC performance achieved by EFB is 1.08\% and 1.06\% higher than the performance of ECB and EPB, respectively, as shown in the fourth row of Table \ref{ablaloss}, which demonstrates that the pseudo-label fusion of ECB and EPB provides reliable supervision for EFB. 
Simultaneously, the cross-supervised learning paradigm of EFB, ECB, and EPB strengthens the shared encoder's ability for the tissue region recognition when introducing $\mathcal{L}_{cs12}$, $\mathcal{L}_{cs21}$, and $\mathcal{L}_{EFB}$, separately. For example, in the sixth-row experiment, the introduction of $\mathcal{L}_{EFB}$ concurrently enhances the performance of ECB and EPB.
In addition, the ensemble results of all branches achieves the highest DSC performance on three different tasks as shown in Table \ref{threeclassifier} and the last row of Table \ref{ablaloss}, which improves DSC by 1.89\%, 1.95\% and 0.97\% compared to ECB, EPB, and EFB, respectively. These extensive results further support the necessity and efficacy of the introduction of EFB.
\begin{table}[h]
	\centering
	\caption{Abaltion study of uncertainty guidance. We report the segmentation performance on DSC.}
	\begin{tabular}{@{}cccc@{}}
		\toprule
		$w_i^1$ & $w_i^2$ & ECB & EPB \\ \midrule
		w/o  & w/o  &    87.97           & 87.11              \\
		w/o  & w    & \textcolor{blue}{89.14}              & 88.59              \\
		w    & w/o  & 88.48              & \textcolor{blue}{89.12}               \\
		w    & w    &\textcolor{red}{89.26}               &\textcolor{red}{89.20}               \\ \bottomrule
	\end{tabular}
	\label{uncertaintygui}
\end{table}

\subsubsection{Ablation study of uncertainty guidance}
\begin{figure}[!t]
	\centering
	\includegraphics[width=3.4in]{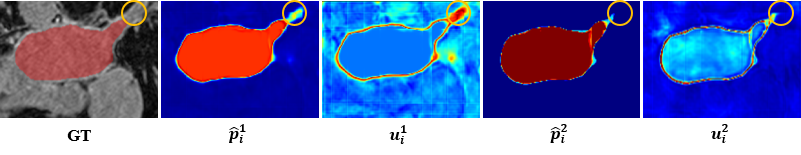}
	\caption{The uncertainty visualization of $\hat{p}_i^1$ and $\hat{p}_i^2$. The proposed evidential uncertainty method shows high-response in error-predicted regions.}
	\label{uncertaintyvis}
\end{figure}

In Fig. \ref{uncertaintyvis}, we visualize the uncertainty maps $u_i^1$ and $u_i^2$ for $\hat{p}_i^1$ and $\hat{p}_i^2$, respectively. As shown in Fig. \ref{uncertaintyvis}, within the yellow circles, both $\hat{p}_i^1$ and $\hat{p}_i^2$ produce false positive predictions. Correspondingly, $u_i^1$ and $u_i^2$ generate high responses in these regions. Thereby, the corresponding weightings $w_i^1$ and $w_i^2$ are small. Consequently, during subsequent training, the contributions of these regions to the overall loss can be reduced to a significant extent through per-pixel weighting using Eq. \ref{eq16} and Eq. \ref{eq17}, thus mitigating the error accumulation issue in semi-supervised training.

Furthermore, we present ablation experiment results in Table \ref{uncertaintygui} with and without uncertainty guidance. The results indicate that introducing both $w_i^1$ and $w_i^2$ simultaneously yields optimal results as shown in the fourth row. Compared to the first row where neither $w_i^1$ nor $w_i^2$ is introduced, ECB generates an improvement of  1.29\% on DSC, and EPB generates an improvement of 2.09\% on DSC.
\subsubsection{Segmentation results under different splits on LA dataset}
\begin{figure}[h]
	\centering
	\includegraphics[width=3.0in]{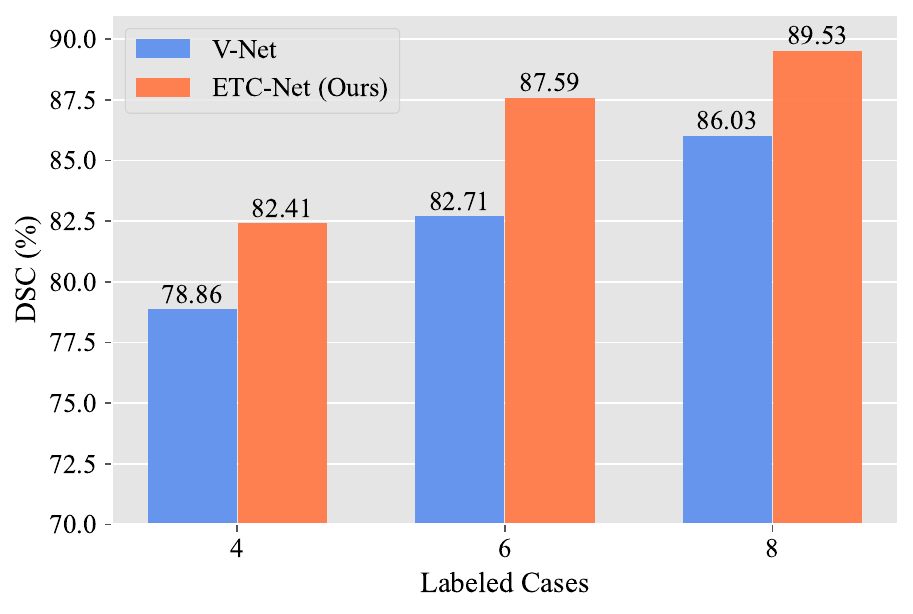}
	\caption{Semi-supervised segmentation results of ETC-Net under different data settings on LA dataset, compared to the supervised trained V-Net only with 4, 6, and 8 labeled cases.}
	\label{differentsplit}
\end{figure}
In Fig. \ref{differentsplit}, we present a comparison of the performance between the semi-supervised ETC-Net using 4, 6, and 8 labeled samples along with the remaining unlabeled samples and the supervised V-Net only using labeled samples. ETC-Net leads to a stable DSC improvement of 3.55\%, 4.88\%, and 3.50\%, which validates the effectiveness of our approach effectiveness in extracting valuable information from unlabeled data by enhancement of cognitive diversity and pseudo-label reliability using the proposed complementary branches. ETC-Net still leads to improved segmentation performance even in scenarios with limited 4 labeled samples.
\section{Conclusion}
\label{Section5}
In this paper, to address issues of cognitive bias and noise components in pseudo labels, we propose a novel evidence-based Tri-branch consistency learning framework, which integrates evidence-based optimization, uncertainty guidance, and evidential decision fusion into ETC-Net, enabling the extraction of valuable information from unlabeled data.
The proposed ECB and EPB facilitate evidential and complementary semi-supervised training in low annotation scenarios, preserving cognitive diversity. Additionally, the uncertain-guided bidirectional cross-supervised training method between ECB and EPB enables beneficial information exchange between the two branches. Furthermore, the introduction of evidential fusion branch supervised by the fused pseudo labels based on Dempster-Shafer theory improves the knowledge distillation from ECB and EPB, further enhancing the semi-supervised training efficiency. The ensemble results of triple branches further improves the decision robustness for semi-supervised medical image segmentation.  Extensive experiments across three benchmarks have demonstrated the effectiveness of the proposed ETC-Net for semi-supervised medical image segmentation tasks. Our future work will focus on studying how to generate substantial and appropriate prediction disagreement on unlabeled data using a more simple way without the introduction of additional structures, such as the exploration on prototypical learning or the efforts on general and simple image augmentation methods.

\scriptsize
\bibliographystyle{IEEEtran}
\bibliography{refs.bib}

\begin{thebibliography}{10}
\providecommand{\url}[1]{#1}
\csname url@samestyle\endcsname
\providecommand{\newblock}{\relax}
\providecommand{\bibinfo}[2]{#2}
\providecommand{\BIBentrySTDinterwordspacing}{\spaceskip=0pt\relax}
\providecommand{\BIBentryALTinterwordstretchfactor}{4}
\providecommand{\BIBentryALTinterwordspacing}{\spaceskip=\fontdimen2\font plus
\BIBentryALTinterwordstretchfactor\fontdimen3\font minus
  \fontdimen4\font\relax}
\providecommand{\BIBforeignlanguage}[2]{{%
\expandafter\ifx\csname l@#1\endcsname\relax
\typeout{** WARNING: IEEEtran.bst: No hyphenation pattern has been}%
\typeout{** loaded for the language `#1'. Using the pattern for}%
\typeout{** the default language instead.}%
\else
\language=\csname l@#1\endcsname
\fi
#2}}
\providecommand{\BIBdecl}{\relax}
\BIBdecl

\bibitem{wang2022disentangled}
J.~Wang, C.~Zhong, C.~Feng, Y.~Zhang, J.~Sun, and Y.~Yokota, ``Disentangled
  representation for cross-domain medical image segmentation,'' \emph{IEEE
  Transactions on Instrumentation and Measurement}, vol.~72, pp. 1--15, 2022.

\bibitem{lin2022ds}
A.~Lin, B.~Chen, J.~Xu, Z.~Zhang, G.~Lu, and D.~Zhang, ``Ds-transunet: Dual
  swin transformer u-net for medical image segmentation,'' \emph{IEEE
  Transactions on Instrumentation and Measurement}, vol.~71, pp. 1--15, 2022.

\bibitem{painchaud2020cardiac}
N.~Painchaud, Y.~Skandarani, Judge, and et~al, ``Cardiac segmentation with
  strong anatomical guarantees,'' \emph{IEEE Transactions on Medical IFmaging},
  vol.~39, no.~11, pp. 3703--3713, 2020.

\bibitem{xu2023cross}
Y.~Xu, K.~Feng, X.~Yan, X.~Sheng, B.~Sun, Z.~Liu, and R.~Yan, ``Cross-modal
  fusion convolutional neural networks with online soft label training strategy
  for mechanical fault diagnosis,'' \emph{IEEE Transactions on Industrial
  Informatics}, 2023.

\bibitem{zhao2019deep}
R.~Zhao, R.~Yan, Z.~Chen, K.~Mao, P.~Wang, and R.~X. Gao, ``Deep learning and
  its applications to machine health monitoring,'' \emph{Mechanical Systems and
  Signal Processing}, vol. 115, pp. 213--237, 2019.

\bibitem{feng2020cpfnet}
S.~Feng and et~al, ``Cpfnet: Context pyramid fusion network for medical image
  segmentation,'' \emph{IEEE Transactions on Medical Imaging}, vol.~39, no.~10,
  pp. 3008--3018, 2020.

\bibitem{cheplygina2019not}
V.~Cheplygina and et~al, ``Not-so-supervised: a survey of semi-supervised,
  multi-instance, and transfer learning in medical image analysis,''
  \emph{Medical Image Analysis}, vol.~54, pp. 280--296, 2019.

\bibitem{tarvainen2017mean}
A.~Tarvainen and H.~Valpola, ``Mean teachers are better role models:
  Weight-averaged consistency targets improve semi-supervised deep learning
  results,'' \emph{Advances in Neural Information Processing Systems}, vol.~30,
  2017.

\bibitem{luo2021efficient}
X.~Luo, W.~Liao, J.~Chen, T.~Song, G.~Wang, S.~Zhang, and et~al, ``Efficient
  semi-supervised gross target volume of nasopharyngeal carcinoma segmentation
  via uncertainty rectified pyramid consistency,'' in \emph{MICCAI}.\hskip 1em
  plus 0.5em minus 0.4em\relax Springer, 2021, pp. 318--329.

\bibitem{ouali2020semi}
Y.~Ouali, C.~Hudelot, and M.~Tami, ``Semi-supervised semantic segmentation with
  cross-consistency training,'' in \emph{CVPR}, 2020, pp. 12\,674--12\,684.

\bibitem{yang2022st++}
L.~Yang, W.~Zhuo, L.~Qi, Y.~Shi, and Y.~Gao, ``St++: Make self-training work
  better for semi-supervised semantic segmentation,'' in \emph{Proceedings of
  the IEEE/CVF Conference on Computer Vision and Pattern Recognition}, 2022,
  pp. 4268--4277.

\bibitem{arazo2020pseudo}
E.~Arazo, D.~Ortego, P.~Albert, N.~E. O’Connor, and K.~McGuinness,
  ``Pseudo-labeling and confirmation bias in deep semi-supervised learning,''
  in \emph{International Joint Conference on Neural Networks}.\hskip 1em plus
  0.5em minus 0.4em\relax IEEE, 2020, pp. 1--8.

\bibitem{wang2023mcf}
Y.~Wang, B.~Xiao, X.~Bi, W.~Li, and X.~Gao, ``Mcf: Mutual correction framework
  for semi-supervised medical image segmentation,'' in \emph{CVPR}, 2023, pp.
  15\,651--15\,660.

\bibitem{chen2021semi}
X.~Chen, Y.~Yuan, G.~Zeng, and J.~Wang, ``Semi-supervised semantic segmentation
  with cross pseudo supervision,'' in \emph{CVPR}, 2021, pp. 2613--2622.

\bibitem{wu2022mutual}
Y.~Wu, Z.~Ge, D.~Zhang, M.~Xu, L.~Zhang, Y.~Xia, and J.~Cai, ``Mutual
  consistency learning for semi-supervised medical image segmentation,''
  \emph{Medical Image Analysis}, vol.~81, p. 102530, 2022.

\bibitem{shi2021inconsistency}
Y.~Shi, J.~Zhang, and et~al, ``Inconsistency-aware uncertainty estimation for
  semi-supervised medical image segmentation,'' \emph{IEEE Transactions on
  Medical Imaging}, vol.~41, no.~3, pp. 608--620, 2021.

\bibitem{yu2019uncertainty}
L.~Yu, S.~Wang, X.~Li, C.-W. Fu, and P.-A. Heng, ``Uncertainty-aware
  self-ensembling model for semi-supervised 3d left atrium segmentation,'' in
  \emph{MICCAI}.\hskip 1em plus 0.5em minus 0.4em\relax Springer, 2019, pp.
  605--613.

\bibitem{li2020self}
Y.~Li, J.~Chen, X.~Xie, K.~Ma, and Y.~Zheng, ``Self-loop uncertainty: A novel
  pseudo-label for semi-supervised medical image segmentation,'' in
  \emph{MICCAI}.\hskip 1em plus 0.5em minus 0.4em\relax Springer, 2020, pp.
  614--623.

\bibitem{he2021re}
R.~He, J.~Yang, and X.~Qi, ``Re-distributing biased pseudo labels for
  semi-supervised semantic segmentation: A baseline investigation,'' in
  \emph{ICCV}, 2021, pp. 6930--6940.

\bibitem{zou2020pseudoseg}
Y.~Zou, Z.~Zhang, H.~Zhang, C.-L. Li, X.~Bian, J.-B. Huang, and T.~Pfister,
  ``Pseudoseg: Designing pseudo labels for semantic segmentation,'' in
  \emph{International Conference on Learning Representations}, 2020.

\bibitem{mittal2019semi}
S.~Mittal and et~al, ``Semi-supervised semantic segmentation with high-and
  low-level consistency,'' \emph{IEEE Transactions on Pattern Analysis and
  Machine Intelligence}, vol.~43, no.~4, pp. 1369--1379, 2019.

\bibitem{li2020shape}
S.~Li, C.~Zhang, and X.~He, ``Shape-aware semi-supervised 3d semantic
  segmentation for medical images,'' in \emph{MICCAI}.\hskip 1em plus 0.5em
  minus 0.4em\relax Springer, 2020, pp. 552--561.

\bibitem{wang2020double}
Y.~Wang, Y.~Zhang, J.~Tian, C.~Zhong, Z.~Shi, Y.~Zhang, and Z.~He,
  ``Double-uncertainty weighted method for semi-supervised learning,'' in
  \emph{MICCAI}.\hskip 1em plus 0.5em minus 0.4em\relax Springer, 2020, pp.
  542--551.

\bibitem{li2020transformation}
X.~Li, L.~Yu, H.~Chen, C.-W. Fu, L.~Xing, and P.-A. Heng,
  ``Transformation-consistent self-ensembling model for semisupervised medical
  image segmentation,'' \emph{IEEE Transactions on Neural Networks and Learning
  Systems}, vol.~32, no.~2, pp. 523--534, 2020.

\bibitem{lai2021semi}
X.~Lai, Z.~Tian, L.~Jiang, S.~Liu, H.~Zhao, L.~Wang, and J.~Jia,
  ``Semi-supervised semantic segmentation with directional context-aware
  consistency,'' in \emph{CVPR}, 2021, pp. 1205--1214.

\bibitem{fan2022ucc}
J.~Fan, B.~Gao, H.~Jin, and L.~Jiang, ``Ucc: Uncertainty guided cross-head
  co-training for semi-supervised semantic segmentation,'' in \emph{CVPR},
  2022, pp. 9947--9956.

\bibitem{li2022trustworthy}
B.~Li, Z.~Han, H.~Li, H.~Fu, and C.~Zhang, ``Trustworthy long-tailed
  classification,'' in \emph{Proceedings of the IEEE/CVF Conference on Computer
  Vision and Pattern Recognition}, 2022, pp. 6970--6979.

\bibitem{dempster1968generalization}
A.~P. Dempster, ``A generalization of bayesian inference,'' \emph{Journal of
  the Royal Statistical Society: Series B (Methodological)}, vol.~30, no.~2,
  pp. 205--232, 1968.

\bibitem{denoeux2000neural}
T.~Denoeux, ``A neural network classifier based on dempster-shafer theory,''
  \emph{IEEE Transactions on Systems, Man, and Cybernetics-Part A: Systems and
  Humans}, vol.~30, no.~2, pp. 131--150, 2000.

\bibitem{huang2021belief}
L.~Huang and et~al, ``Belief function-based semi-supervised learning for brain
  tumor segmentation,'' in \emph{IEEE 18th International Symposium on
  Biomedical Imaging}.\hskip 1em plus 0.5em minus 0.4em\relax IEEE, 2021, pp.
  160--164.

\bibitem{huang2023semi}
L.~Huang, S.~Ruan, and T.~Den{\oe}ux, ``Semi-supervised multiple evidence
  fusion for brain tumor segmentation,'' \emph{Neurocomputing}, 2023.

\bibitem{huang2022evidence}
L.~Huang, T.~Denoeux, P.~Vera, and S.~Ruan, ``Evidence fusion with contextual
  discounting for multi-modality medical image segmentation,'' in
  \emph{MICCAI}.\hskip 1em plus 0.5em minus 0.4em\relax Springer, 2022, pp.
  401--411.

\bibitem{diao2021efnet}
Z.~Diao, H.~Jiang, X.-H. Han, Y.-D. Yao, and T.~Shi, ``Efnet: evidence fusion
  network for tumor segmentation from pet-ct volumes,'' \emph{Physics in
  Medicine \& Biology}, vol.~66, no.~20, p. 205005, 2021.

\bibitem{kendall2017uncertainties}
A.~Kendall and Y.~Gal, ``What uncertainties do we need in bayesian deep
  learning for computer vision?'' \emph{Advances in Neural Information
  Processing Systems}, vol.~30, 2017.

\bibitem{xie2016melanoma}
F.~Xie and et~al, ``Melanoma classification on dermoscopy images using a neural
  network ensemble model,'' \emph{IEEE Transactions on Medical Imaging},
  vol.~36, no.~3, pp. 849--858, 2016.

\bibitem{tong2021evidential}
Z.~Tong, P.~Xu, and T.~Denoeux, ``Evidential fully convolutional network for
  semantic segmentation,'' \emph{Applied Intelligence}, vol.~51, pp.
  6376--6399, 2021.

\bibitem{sensoy2018evidential}
M.~Sensoy, L.~Kaplan, and M.~Kandemir, ``Evidential deep learning to quantify
  classification uncertainty,'' \emph{Advances in Neural Information Processing
  Systems}, vol.~31, 2018.

\bibitem{zou2022tbrats}
K.~Zou, X.~Yuan, X.~Shen, M.~Wang, and H.~Fu, ``Tbrats: Trusted brain tumor
  segmentation,'' in \emph{MICCAI}.\hskip 1em plus 0.5em minus 0.4em\relax
  Springer, 2022, pp. 503--513.

\bibitem{li2022region}
H.~Li, Y.~Nan, J.~Del~Ser, and G.~Yang, ``Region-based evidential deep learning
  to quantify uncertainty and improve robustness of brain tumor segmentation,''
  \emph{Neural Computing and Applications}, pp. 1--15, 2022.

\bibitem{zhang2018generalized}
Z.~Zhang and M.~Sabuncu, ``Generalized cross entropy loss for training deep
  neural networks with noisy labels,'' \emph{Advances in neural information
  processing systems}, vol.~31, 2018.

\bibitem{xiong2021global}
Z.~Xiong and et~al, ``A global benchmark of algorithms for segmenting the left
  atrium from late gadolinium-enhanced cardiac magnetic resonance imaging,''
  \emph{Medical Image Analysis}, vol.~67, p. 101832, 2021.

\bibitem{clark2013cancer}
K.~Clark, B.~Vendt, and et~al, ``The cancer imaging archive (tcia): maintaining
  and operating a public information repository,'' \emph{Journal of Digital
  Imaging}, vol.~26, no.~6, pp. 1045--1057, 2013.

\bibitem{bernard2018deep}
O.~Bernard and et~al, ``Deep learning techniques for automatic mri cardiac
  multi-structures segmentation and diagnosis: is the problem solved?''
  \emph{IEEE Transactions on Medical Imaging}, vol.~37, no.~11, pp. 2514--2525,
  2018.

\bibitem{luo2021semi}
X.~Luo, G.~Wang, and et~al, ``Semi-supervised medical image segmentation
  through dual-task consistency,'' in \emph{Proceedings of the AAAI Conference
  on Artificial Intelligence}, vol.~35, no.~10, 2021, pp. 8801--8809.

\bibitem{9966841}
T.~Lei, D.~Zhang, X.~Du, X.~Wang, Y.~Wan, and A.~K. Nandi, ``Semi-supervised
  medical image segmentation using adversarial consistency learning and dynamic
  convolution network,'' \emph{IEEE Transactions on Medical Imaging}, vol.~42,
  no.~5, pp. 1265--1277, 2023.

\end{thebibliography}

\end{document}